\documentclass{article} 
\usepackage{iclr2027_conference,times}

\usepackage{amsmath,amsfonts,bm}

\def\eqref#1{equation~\ref{#1}}

\def\1{\bm{1}}

\DeclareMathAlphabet{\mathsfit}{\encodingdefault}{\sfdefault}{m}{sl}
\SetMathAlphabet{\mathsfit}{bold}{\encodingdefault}{\sfdefault}{bx}{n}

\usepackage{hyperref}
\usepackage{url}

\usepackage{booktabs}
\usepackage{xcolor}
\usepackage{amsthm}
\usepackage{textgreek}
\usepackage{wrapfig}
\definecolor{dark green}{rgb}{0.0,0.5,0.0}

\newcommand{\rev}[1]{\textcolor{black}{#1}} 

\usepackage{algorithm}
\usepackage{algorithmic}
\usepackage{comment}
\usepackage{bbm}

\usepackage{graphicx}
\usepackage{subcaption}
\usepackage{tikz}
\usepackage{pgfplots}
\pgfplotsset{compat=1.18}

\title{Pitfalls of Dependency-Ready Decoding for Masked Diffusion Language Models}

\author{Lina Berrayana \\ NEC Corporation, Tokyo, Japan \\ EPFL, Lausanne, Switzerland \\
\texttt{lina.berrayana@alumni.epfl.ch}
\And
Christopher Malon \\ NEC Corporation, Tokyo, Japan \\ NEC Laboratories America, Princeton, USA \\
\texttt{christopher.malon@nec.com}
\And
Yoichi Ishibashi \\ NEC Corporation, Tokyo, Japan \\
\texttt{yoichi-ishibashi@nec.com}
\AND
Taichi Aida \\ NEC Corporation, Tokyo, Japan \\
\texttt{taichi-aida@nec.com}
\And
Masafumi Oyamada \\ NEC Corporation, Tokyo, Japan \\
\texttt{oyamada@nec.com}
}

\newcommand{\Fut}{\mathcal{F}}

\newcommand{\norm}[1]{\vert\vert#1\vert\vert}
\newcommand{\abs}[1]{\vert#1\vert}

\iclrfinalcopy 
\begin{document}

\maketitle

\begin{abstract}
Masked diffusion models (MDMs) have shown strong performance
on structured reasoning tasks such as Sudoku, Trip Planning,
and Countdown. 
MDMs introduce a freedom in the order in which
tokens may be unmasked, and to exercise this freedom for good
accuracy, early decoding strategies relied on signals of
confidence or the downstream impact of decoding a token.
However, simple examples seem to illustrate that resolving
a token's dependencies before unmasking it should be important.
Therefore, we examine whether this principle, which we call \emph{dependency-ready decoding}, would
contribute to better accuracy, from a practical and a theoretical
perspective.  We examine a recent decoder that implements
dependency-readiness through attention, and generally find that when it performs better than confidence-based or future utility baselines,
its decoding order collapses to left-to-right decoding.
For theoretical analysis, we establish a definition of
dependency without resorting to attention, measured
by an information-theoretic score of how much revealing a single masked token (or a set) would shift the model’s prediction for another. We prove that when the margin is large relative to this score,
no such revelation can change the top prediction.
These findings leave only a narrow scope for future dependency-ready decoders: their potential benefit is limited to positions with margins smaller than the information-theoretic sensitivity, and reliance on attention should be questioned. 
\end{abstract}

\section{Introduction}
\label{sec:intro}

Diffusion models have achieved remarkable success in image generation, and
recent work has extended them to language. Masked diffusion language models
(MDMs) \citep{DBLP:conf/nips/AustinJHTB21,DBLP:conf/icml/LouME24,DBLP:conf/nips/SahooASGMCRK24,DBLP:conf/nips/ShiHWDT24}, the
dominant approach in this setting, generate text by iteratively revealing
tokens from a fully masked sequence, rather than following the fixed
left-to-right factorization of autoregressive models. At inference time, this non-causal generation process offers two key advantages. First, multiple tokens can be
committed in parallel during each denoising step, substantially improving
decoding throughput
\citep{DBLP:journals/corr/abs-2506-17298,gemini2025diffusion,DBLP:journals/corr/abs-2505-22618, DBLP:conf/iclr/ArriolaGCYQHSK25, DBLP:conf/nips/Ben-HamuGSNK25, dapd2026}.
Second, the model is free to choose the order in which tokens are revealed.
MDMs have shown strong performance on code generation
\citep{DBLP:journals/corr/abs-2506-20639,DBLP:journals/corr/abs-2509-01142}
and, most notably, on structured reasoning tasks such as Sudoku
\citep{DBLP:conf/nips/ShahDWP24,DBLP:conf/icml/KimSKKC25}, Trip Planning
\citep{DBLP:journals/corr/abs-2508-15487}, and Countdown
\citep{DBLP:conf/iclr/YeGGZJLK25}. Prior work attributes this success to their
flexible generation order \citep{DBLP:conf/icml/KimSKKC25,DBLP:conf/iclr/YeGGZJLK25}, which may allow decoding to better accommodate the
underlying dependency structure of these tasks
\citep{DBLP:conf/iclr/YeGGZJLK25}. With the emergence of large
instruction-tuned MDMs such as LLaDA \citep{DBLP:conf/nips/NieZYZOHZLWL25} and Dream
\citep{DBLP:journals/corr/abs-2508-15487}, the decoding policy has become a
central component of inference.

MDMs generate text through iterative denoising: at each step, the
model predicts all masked tokens, and a decoding method selects a
subset to reveal, which then serves as context for later steps.
While the first MDMs revealed positions at
random~\citep{DBLP:conf/nips/SahooASGMCRK24}, later methods relied
on confidence-based criteria such as maximum
probability~\citep{DBLP:conf/nips/NieZYZOHZLWL25}, maximum
margin~\citep{DBLP:conf/icml/KimSKKC25}, or minimum
entropy~\citep{DBLP:journals/corr/abs-2508-15487}. More recent
methods, such as the Info-Gain Sampler~\citep{yang2026improving},
SOAR~\citep{DBLP:journals/corr/abs-2602-10953}, and
DAPD~\citep{dapd2026}, instead focus on \emph{future utility},
that is, how much revealing a token helps predict the tokens that
are still masked. 

Considering simple puzzle examples such as Sudoku suggests
that resolving token dependencies before unmasking a token
might be important, an intuition which we explore in
Section~\ref{sec:motivating_dependency}.  
This principle, which we call 
{\em dependency-ready decoding}, at first appears to be
different from the confidence or future utility considerations.
The recent Dependency-Oriented Sampler (DOS)~\citep{DBLP:conf/acl/ZhouHH26} follows this principle using attention
as a proxy for dependency, following 
DAPD~\citep{dapd2026} and
ADAS~\citep{DBLP:journals/corr/abs-2606-10829},
which had used attention to avoid parallel decoding conflicts.
At each step, DOS reveals the masked positions with the highest
\emph{groundedness}, that is, the ones that attend most to
the tokens already revealed.

In this paper, we describe some pitfalls that implementations of
dependency-ready decoding may confront, both practically in an existing
attention-based implementation (DOS), and in theory more generally.

From a practical perspective, we reimplement DOS on LLaDA and Dream and add a left-to-right baseline that prior work did not report.  We find that generally, when DOS performs better than confidence-based or future utility baselines, its
decoding order collapses to left-to-right decoding.
Its previously reported
advantage over confidence-based methods shrinks or disappears once the
end-of-sequence (EOS) fix of \citet{DBLP:conf/nips/NieZYZOHZLWL25} is applied.
We show why the left-to-right collapse should be expected because of the use
of attention at the first layer as a proxy for dependency, but empirically,
we observe a collapse even if attention is taken at late layers.

From a theoretical perspective, we show conditions under which
Maximum-Margin decoding already follows a dependency-ready principle.
Precisely, we introduce the Pre-Commit score as a measure of dependency readiness,
which is an information-theoretic measure of how far the prediction for a masked
token can shift when any
single other masked token (or a set of tokens) is revealed at its ground-truth value.
This score does not rely on attention as a proxy for dependency, although
it is useful only for theoretical analysis because it requires ground truth
token values.
We prove that when the margin is large relative to this score, no revelation of one or a set of tokens can change the top prediction, so Maximum-Margin decoding already
commits dependency-ready tokens in these cases. \rev{ We then measure, on Sudoku,
how often the commits of each decoder satisfy this condition. } 

Together, these results suggest that dependency-ready decoding
strategies should reconsider their reliance on attention,
and that their benefit may be limited to decoding token
positions with margins that are small with respect to the Pre-Commit score\rev{, which still make up a substantial
share of Maximum-Margin commits}.


In summary, our contributions are:
\begin{enumerate}
\item We highlight and explain a collapse of attention-based dependency-ready decoding to left-to-right decoding. 
\item We show that an EOS handling issue  confounds previous comparisons of dependency-ready decoding strategies against confidence decoding strategies.
\item We establish a Pre-Commit score as a possible definition of dependency readiness, in an oracle setting that does not use attention.
\item 
\rev{We prove that a large margin relative to the Pre-Commit score certifies that revealing masked tokens will not change a prediction.}
\item \rev{We measure how often existing decoders commit tokens that satisfy the Pre-Commit criterion, and how often those predictions are accurate.}

\end{enumerate}

\section{Background}
\label{sec:background}

\paragraph{Masked diffusion language models.}
MDMs \citep{DBLP:conf/nips/AustinJHTB21,DBLP:conf/icml/LouME24,DBLP:conf/nips/SahooASGMCRK24,DBLP:conf/nips/ShiHWDT24}
generate a sequence by iteratively replacing \textsc{[MASK]} tokens with
vocabulary symbols. Let $\mathbf{r}$ denote the prompt and
$\mathbf{x}^{(t)}\in(\mathcal{V}\cup\{\mathbf{m}\})^L$ the partially decoded
sequence at reverse step $t$. The remaining masked positions are
$\Fut_t=\{v:x_v^{(t)}=\mathbf{m}\}$, while the revealed context consists of
the prompt together with every committed token,
$\mathcal R_t=(\mathbf{r},\{x_v^{(t)}\}_{v\notin\Fut_t})$. At every reverse
step the model predicts a categorical distribution
$p_\theta(X_v\mid\mathcal R_t)$ for every remaining masked position
$v\in\Fut_t$, from which the decoder selects positions to reveal.

\paragraph{Decoding as an ordering problem.}
Unlike autoregressive models, which fix a left-to-right factorization, an MDM
is free to choose which positions to commit at each reverse step. A decoding
policy therefore maps the current state to a subset of masked positions,
$\pi:(\mathcal R_t,\Fut_t)\rightarrow S_t\subseteq\Fut_t$. When $|S_t|=1$,
the policy reduces to a permutation of the generation positions. We refer to
this permutation as the \emph{decoding order}. Existing decoders differ almost
entirely in how they choose this order.
Practical MDMs often commit multiple positions simultaneously for efficiency.
This approximates the joint predictive distribution as
\begin{equation}
p_\theta(X_{S_t}\mid\mathcal R_t)\approx\prod\nolimits_{v\in S_t}p_\theta(X_v\mid\mathcal R_t),
\label{eq:factorization}
\end{equation}
which is exact only if the selected positions are conditionally independent.
Consequently, decoding quality depends not only on prediction accuracy but also
on whether positions are revealed in an order consistent with their underlying
dependencies.

\section{Why Use Dependency Information in Decoding?}
\label{sec:motivating_dependency}

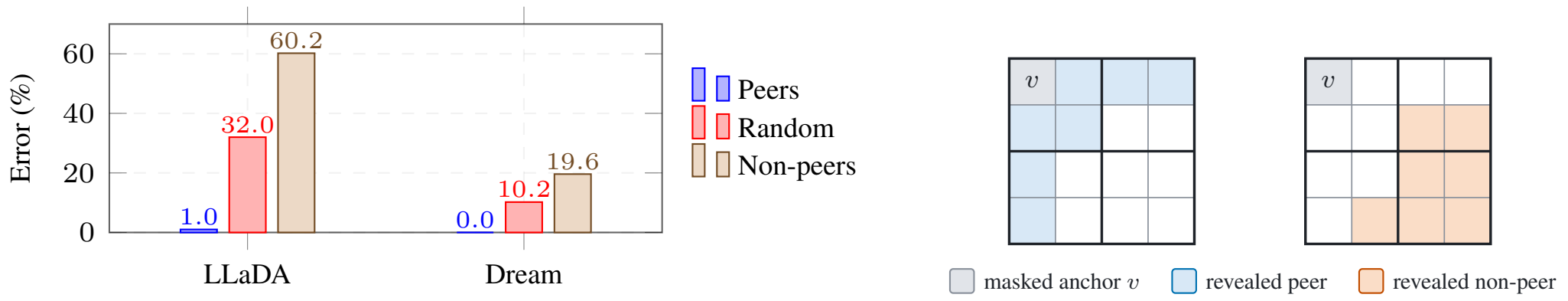
\begin{figure}[h]
    \centering
    \captionsetup[subfigure]{skip=2pt}
    \captionsetup{skip=4pt}
    \vspace{-5mm}

    \begin{subfigure}[t]{0.59\linewidth}
        \centering
        \begin{tikzpicture}
        \begin{axis}[
            scale only axis,
            width=0.58\linewidth,
            height=1.8cm,
            ybar=3pt,
            bar width=9pt,
            ymin=0,
            ymax=70,
            ytick={0,20,40,60},
            ylabel={Error (\%)},
            symbolic x coords={LLaDA,Dream},
            xtick=data,
            x tick label style={font=\scriptsize},
            yticklabel style={font=\scriptsize},
            ylabel style={font=\scriptsize},
            tick label style={font=\scriptsize},
            enlarge x limits=0.5,
            grid=major,
            grid style={
                dashed,
                opacity=0.20
            },
            axis line style={opacity=0.6},
            legend style={
                draw=none,
                font=\scriptsize,
                at={(1.03,0.5)},
                anchor=west,
                row sep=-2pt
            },
            legend cell align=left,
            nodes near coords,
            nodes near coords style={
                font=\tiny,
                /pgf/number format/fixed,
                /pgf/number format/fixed zerofill,
                /pgf/number format/precision=1,
                inner sep=1pt
            },
            clip=false
        ]

        \addplot coordinates {(LLaDA,1.0) (Dream,0.0)};
        \addlegendentry{Peers}

        \addplot coordinates {(LLaDA,32.0) (Dream,10.2)};
        \addlegendentry{Random}

        \addplot coordinates {(LLaDA,60.2) (Dream,19.6)};
        \addlegendentry{Non-peers}

        \end{axis}
        \end{tikzpicture}

        \caption{
        }
        \label{fig:peer_random_nonpeer}
    \end{subfigure}
    \hfill
    \begin{subfigure}[t]{0.40\linewidth}
        \centering
        \includegraphics[
            width=\linewidth,
            height=2.1cm,
            keepaspectratio
        ]{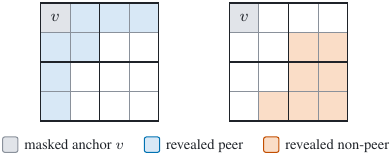}

        \caption{
        }
        \label{fig:sudoku_peers}
    \end{subfigure}
    \caption{(a) \textbf{Context relevance matters.} Prediction error at a masked target cell after revealing seven of its peers, seven random cells, or seven non-peers, over 120 $4{\times}4$ Sudoku puzzles. (b) The seven peers (left) and seven non-peers (right) of a target cell $v$.
    }
\end{figure}

In the $4\times4$ Sudoku task of \citet{DBLP:conf/iclr/YeGGZJLK25},
each puzzle has a unique solution, and the rules make
explicit which cells depend on which. For a target cell
$v$, we call the cells in its row, column, or $2\times2$
subgrid its \emph{peers} (Figure~\ref{fig:sudoku_peers}).
Peers directly constrain the value of $v$, while the
remaining cells, the \emph{non-peers}, do not.
To separate which cells are revealed from how many are
revealed, we start from the same partially revealed puzzle,
reveal seven cells at their ground-truth values, and then
measure the prediction error at a fixed target cell. The
seven cells are all peers, all non-peers, or chosen at
random. Some of them may already be revealed, in which case
fewer than seven are new.
Figure~\ref{fig:peer_random_nonpeer} shows the results for
LLaDA and Dream. With the peers revealed, both models make
almost no errors on the target (1.0\% for LLaDA and 0.0\%
for Dream). With the non-peers revealed, the error reaches
60.2\% and 19.6\%, and random cells fall in between. The
same amount of context therefore helps much more when it
contains the target's peers.

\begin{figure}[h]
  \centering
  \includegraphics[width=\textwidth]{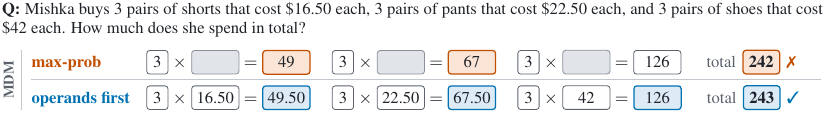}
  \caption{GSM8K example: the same MDM decoded in two orders.}
  \label{fig:gsm8k_example}
\end{figure}

We observe the same phenomenon in individual GSM8K
\citep{DBLP:journals/corr/abs-2110-14168} examples
(Figure~\ref{fig:gsm8k_example}). In one example, computing
a total cost requires intermediate products such as
$3\times16.50$ and $3\times22.50$. Under maximum-probability
decoding, the model commits the result tokens before the
operand tokens are revealed, so it must estimate the
intermediate values from incomplete context. This gives $49$
instead of $49.50$ and $67$ instead of $67.50$, and a wrong
total of $242$. We then repeat the decoding but forbid
selecting a result token while any of its operand tokens is
still masked. With this constraint, the model produces the
exact intermediate values $49.50$, $67.50$, and $126$, and
the correct total of $243$. The model, the
maximum-probability rule, and the decoding budget stay the
same, so only the decoding order differs.

In both cases, the model predicts a token more accurately once specific other tokens are revealed, namely the peers of a Sudoku cell or the operands of a GSM8K result. This motivates dependency-ready decoding, which delays committing a token until most of the tokens that affect its prediction are revealed, a notion we make precise in Section~\ref{sec:without_attention}. A task-agnostic decoder, however, does not know these tokens in advance and has to estimate them from the model. In Section~\ref{sec:exp}, we study the use of attention to estimate these tokens, as proposed in DOS.

\section{Related Work}
\label{sec:relatedwork}

\paragraph{Confidence-based decoding.}
Early MDMs selected decoding positions uniformly at random, mirroring the masking process used during training
\citep{DBLP:conf/nips/AustinJHTB21,DBLP:conf/icml/LouME24,DBLP:conf/nips/SahooASGMCRK24}.
As larger MDMs emerged, confidence-based scheduling became the dominant decoding strategy. Variations on confidence-based scheduling rank masked positions according to maximum token probability 
\citep{DBLP:conf/nips/NieZYZOHZLWL25},
maximum margin \citep{DBLP:conf/icml/KimSKKC25}, or predictive entropy \citep{DBLP:conf/nips/Ben-HamuGSNK25}. The most common scores are
\[
s_v^{\mathrm{conf}}=\max_w p_\theta(x_v=w\mid\mathcal R_t),\qquad
s_v^{\mathrm{margin}}=p_\theta(x_v=w_1\mid\mathcal R_t)-p_\theta(x_v=w_2\mid\mathcal R_t),
\]
where $s_v$ denotes the confidence score assigned to the masked position $v$, and $w$ ranges over vocabulary tokens. In the margin score, $w_1$ and $w_2$ denote the most likely and second-most-likely tokens, respectively. Although these heuristics differ in how confidence is measured, they all operate on the same marginal predictive distribution
$p_\theta(X_v\mid\mathcal R_t)$.
Consequently, they estimate how certain the model is about its current prediction, but not whether that prediction still depends on information contained in unrevealed tokens.

\paragraph{Beyond confidence-based decoding.}
Recent work has explored decoding strategies that move beyond purely greedy confidence-based scheduling. We broadly divide these approaches into three categories. The first introduces additional information or supervision to learn or construct a desirable decoding trajectory. For example, \citet{DBLP:journals/corr/abs-2602-09501} learns an unmasking policy from an oracle trajectory, while \citet{DBLP:conf/acl/HuangLLYWXCS26} uses an auxiliary mechanism to guide unmasking. Such approaches can provide richer trajectory-level information, but introduce an additional source of supervision or computation.

The second category takes a \emph{utilitarian} approach, evaluating how a current unmasking decision may affect future decoding. Information-Gain Sampler \citep{yang2026improving} selects tokens based on their expected reduction of marginal entropy in the remaining masked positions, while SOAR \citep{DBLP:journals/corr/abs-2602-10953} searches over alternative unmasking orders when confidence is insufficient. 
Dependency-Aware Parallel Decoding (DAPD) \citep{dapd2026}, despite the word dependency, shares the same principle.  At each step, it selects one or more tokens with the {\em highest impact} on other masked tokens (as measured by symmetrized attention), and decodes them.

A contrary principle is to prioritize the \emph{readiness} of the decoded tokens, rather than future utility.  DOS \citep{DBLP:conf/acl/ZhouHH26} examines whether 
a token to be decoded has unresolved dependencies according to the
attention matrix, {\em i.e.}, whether it is {\em ready} to be decoded.  
It is the focus of our analysis.
For a masked target position $v \in F_t$, it defines the dependency score as groundedness:
\begin{equation}
Gr(v) = \sum_{u \in \mathcal R_t} \operatorname{Attn}(v,u),
\end{equation}
measuring the extent to which the token at position $v$ attends to already {\em unmasked} positions $u$.
In the sequential variant, each iteration step decodes at the position with the
maximum groundedness score.  Because it is score-based, this strategy
is compatible with parallel decoding methods.

Other methods, including KLASS \citep{DBLP:conf/nips/KimHJPY25}, Fast-dLLM \citep{DBLP:journals/corr/abs-2505-22618}, and Prophet \citep{li2026diffusion}, primarily target inference acceleration through more aggressive parallel decoding. These methods reduce the number of denoising steps but do not directly address which positions are safest to commit. In the parallel setting, DAPD avoids simultaneously decoding tokens with strong dependencies to each other, as measured by symmetrized self-attention.  The Entropy-Bound Sampler \citep{DBLP:conf/nips/Ben-HamuGSNK25} can be applied with any score-based decoding strategy to choose multiple tokens to decode at each step, up to an entropy budget.  In this way, the parallel decoding issue may be considered as orthogonal to a score which prioritizes tokens to be decoded.


Block diffusion \citep{DBLP:conf/iclr/ArriolaGCYQHSK25} is another way of achieving parallelization while
avoiding the dangers of predicting faraway tokens.  It constrains blocks to
be predicted from left to right, with multiple tokens selected at each step
within a single block until the block is completed.  In this way, it provides
an autoregressive heuristic to decoding.  Because, as we will show, DOS already
mostly follows a left-to-right order when given complete freedom, there is
no need for us to require block structure, a hyperparameter setting which would
confound a comparison between methods.

\rev{We do not study revocable decoding strategies such as WINO \citep{wino} and DARD \citep{dard} because revocable decisions make it less important to select dependency-ready tokens. Our stability result
still bears on them: a prediction that satisfies the criterion of Section~\ref{sec:which0} keeps its top-1 prediction
after any single further revelation, so it is a prediction that a revocable decoder would not need to
revise. Conversely, revocation can repair the unstable predictions that Section~\ref{sec:which} finds for Maximum-Margin
decoding.}

At the time of writing, DOS implements the purest form of attention-based
dependency-ready decoding.  Other recent dependency-aware decoders \citep{dawn, clad} have
emerged, which use dependency relations estimated by attention scores to avoid
conflicts when tokens are simultaneously decoded, but the prioritization of
tokens to decode is first driven by confidence.  Their mixture of confidence
scores and heuristics with thresholded attention scores introduces confounding
phenomena which we avoid in our analysis.


\section{Dependency-Ready Decoding With Attention}
\label{sec:exp}

Section~\ref{sec:motivating_dependency} shows that a token
is easier to predict once the tokens that constrain it are
revealed.
We now test whether a decoder that uses attention to find
tokens with resolved dependencies improves accuracy in
practice. We focus on DOS described in Section~\ref{sec:relatedwork}, which decodes the masked
position that attends most to already revealed tokens, and
compare it with confidence-based decoders, decoders based
on future utility (Info-Gain and DAPD), and a left-to-right
baseline. For each method, we measure accuracy and how far
its decoding order is from the left-to-right order. Because
DOS was originally compared without the EOS fix described
below, we report results with and without it.

\subsection{Experimental Setup}
Unless otherwise stated, all experiments use
LLaDA-8B-Instruct~\citep{DBLP:conf/nips/NieZYZOHZLWL25} or Dream-v0-Instruct-7B \citep{DBLP:journals/corr/abs-2508-15487} with the same prompts,
denoising schedule, and decoding budget across all methods.
\footnote{\rev{The latest models in the LLaDA series are LLaDA 2.0, 2.1 and 2.2, but LLaDA 2.0 \citep{llada2} was not considered because of its dependence on block diffusion, and LLaDA 2.1 \citep{llada21} and 2.2 \citep{llada22} may revoke previous decoding decisions,
making the notion of dependency readiness less relevant.}}

The only difference between policies is the order in which masked positions
are committed; token predictions are always obtained from the same model
without retraining or additional forward passes.
All experiments were conducted on a compute server equipped with eight NVIDIA
A100-PCIE GPUs (40\,GB memory each) running CUDA~12.2.

\paragraph{The EOS Fix.}
The LLaDA paper \citep{DBLP:conf/nips/NieZYZOHZLWL25} suggested a measure to prevent the premature generation of EOS tokens.  When uncontrolled, models may tend to fill the tail of their generation context with EOS tokens before committing any meaningful tokens, ultimately not leaving enough room for meaningful content \citep{kim2026rainbow}.
This measure, which we refer to as the ``EOS Fix,'' stops the selection of positions where EOS is the maximum-probability token candidate, unless EOS is the maximum-probability token candidate at every masked position.  Effectively, it is a filter on the masked positions that may be filled by a decoding strategy at a given step.
Our results are reported for decoding methods with the EOS Fix either on or off, as indicated.

\paragraph{Tasks.}
Tables~\ref{tab:benchlrllada} and~\ref{tab:benchlrdream} report results on GSM8K~\citep{DBLP:journals/corr/abs-2110-14168} and  MATH500 \citep{DBLP:conf/nips/HendrycksBKABTS21}, covering diverse
multi-step reasoning and structured prediction tasks, on the coding tasks HumanEval~\citep{DBLP:journals/corr/abs-2107-03374} and MBPP \citep{DBLP:journals/corr/abs-2108-07732}, and on the puzzle tasks of Sudoku 9x9 \citep{DBLP:conf/nips/ShahDWP24}, Trip Planning \citep{DBLP:journals/corr/abs-2406-04520}, and Countdown with three numbers \citep{DBLP:conf/iclr/YeGGZJLK25}. For Sudoku 9x9, we measure per-token accuracy.
All remaining benchmarks are evaluated using the official evaluation suite released with Dream \citep{DBLP:journals/corr/abs-2508-15487}, keeping its datasets, prompt templates, in-context exemplars and scoring scripts unchanged.
In all runs, the number of decoding steps equals the generation length. Following Dream's evaluation scripts, we use 256 tokens for GSM8K, 512 for MATH500, 256 for Trip Planning and 24 for Countdown. For HumanEval and MBPP, we replace Dream's 768 and 1024 tokens with LLaDA's 512 and 256~\citep{DBLP:conf/nips/NieZYZOHZLWL25}, closer to the 256 used by DOS~\citep{DBLP:conf/acl/ZhouHH26} and DAPD~\citep{dapd2026}. For Sudoku 9x9, we use 81 tokens, one per cell, as in \citet{DBLP:conf/iclr/YeGGZJLK25}.

\paragraph{Decoding strategies.}
We evaluate maximum margin, maximum probability (``standard''), minimum
entropy, DOS, DAPD, and Info-Gain Sampler against a baseline of left-to-right decoding.  To give the sampler full freedom to decode any token and avoid confounding issues of parallelism, no blocks are used, and each method is evaluated in a setting where a single token is decoded at each step.  The DOS authors found that different layers performed optimally for different tasks, but that the first layer was a consistently good choice \citep{DBLP:conf/acl/ZhouHH26}.  We evaluate this as ``DOS (1).'' DAPD reports that upper-layer attention best recovers a known synthetic Markov Random Field structure, and therefore averages attention over the final 30\% of layers in large models, so we implement DOS with these final layers as well for comparison (``DOS (late)'').

\begin{table}[t]
\centering
\small
\setlength{\tabcolsep}{4pt}
\begin{tabular}{lc|cc|cc|ccc}
\toprule
\textbf{Method} & \textbf{EOS fix}
& \textbf{GSM8K} 
& \textbf{MATH500}
& \textbf{HumanEval}
& \textbf{MBPP} 
& \textbf{Trip}
& \textbf{Sudoku}
& \textbf{cd3}
\\
\midrule
Max Prob & Off & 50.49 & 19.20 & 15.24 & 22.60 & 10.12 & 37.06 &  45.36 \\
Max Margin & Off & 50.34 & 19.40 & 15.85 & 25.40 & 11.75 &  38.15 & 44.76\\
Min Entropy & Off & 52.39 & 20.00 & 23.17 & 16.80 & 6.44 & 35.32 & 40.73 \\
\midrule
Max Prob & On & 63.84 & 27.20 & 28.05 & 39.00 & 10.38 & {\bf 39.60} & 36.29 \\
Max Margin & On & 64.37 & 28.20 & 29.88 & {\bf 42.80} & 9.62 & 39.34 & 37.40 \\
Min Entropy & On & 63.47 & 25.80 & {\bf 37.20} & 40.60 & 9.75 & 39.51 & 36.90 \\
\midrule
DOS (1) & On & 76.95 & {\bf 37.00} & 36.59 & 37.80 & 5.81 & 21.67 & 32.36 \\
DOS (late) & On & {\bf 77.18} & 36.40 & 35.37 & 37.60 & 6.56 & 21.71 & 31.05 \\
Info-Gain & On & 65.88 & 27.40 & 29.27 & 39.60 & 10.25 & 37.90 & 34.38 \\
DAPD & On & 65.05 & 24.40 & 26.83 & 37.40 & 8.69 & 34.58 & 36.39 \\
\midrule
DOS (1) & Off & 76.95 & {\bf 37.00} & 36.59 & 37.80 & 6.69 & 21.71 & 32.36 \\
DOS (late) & Off & 74.45 & 36.20 & 35.37 & 37.80 & 6.62 & 21.40 & 33.27 \\
Info-Gain & Off & 52.08 & 18.80 & 17.68 & 27.00 & {\bf 16.25} & 36.58 & 44.46 \\
DAPD & Off & 53.30 & 21.80 & 11.59 & 11.60 & 4.81 & 32.04 & {\bf47.37} \\
\midrule
Left to Right &   & 76.95 & {\bf 37.00} & 36.59 & 37.80 & 6.69 & 21.71 & 32.36 \\
\bottomrule
\end{tabular}
\caption{Evaluation of decoding methods on LLaDA Instruct, with
temperature 0 and no Block Diffusion, on GSM8K-CoT(8-shot), MATH500(4-shot), HumanEval(0-shot), MBPP(0-shot), Trip Planning (2-shot), 9x9 Sudoku  (5-shot), and 3-number Countdown (8-shot).}

\label{tab:benchlrllada}
\end{table}

\begin{table}[t]
\centering
\small
\setlength{\tabcolsep}{4pt}
\begin{tabular}{lc|cc|cc|ccc}
\toprule
\textbf{Method} & \textbf{EOS fix}
& \textbf{GSM8K}
& \textbf{MATH500}
& \textbf{HumanEval}
& \textbf{MBPP}
& \textbf{Trip}
& \textbf{Sudoku}
& \textbf{cd3}
\\
\midrule
Max Prob & Off & 37.83 & 7.60 & 26.83 & 45.40 & 14.81 & 71.89 & 53.53 \\
Max Margin & Off & 39.73 & 8.20 & 28.66 & 47.40 & 14.25 & {\bf 72.21} & 53.12 \\
Min Entropy & Off & 35.63 & 7.00 & 21.95 & 37.00 & 15.75 & 69.55 & 53.02 \\
\midrule
Max Prob & On & 80.59 & 37.20 & 53.66 & 56.80 & 15.94 & 70.11 & 54.74 \\
Max Margin & On & 81.12 & 36.80 & 46.95 & 55.80 & 15.88 & 70.27 & 52.62 \\
Min Entropy & On & {\bf 82.64} & 40.60 & 55.49 & 56.80 & 15.69 & 69.03 & 53.02  \\
\midrule
DOS (1) & On & 81.05 & {\bf 41.60} & {\bf 58.54} & {\bf 57.60} & 11.88 & 49.40 & 45.26 \\
DOS (late) & On & 77.86 & 33.20 & 53.66 & 56.80 & 13.31 & 49.40 & 45.26 \\
Info-Gain & On & 81.35 & 40.80 & {\bf 58.54} & 56.20 & 15.94 & 70.11 & 54.74 \\
DAPD & On & 73.69 & 32.40 & 45.12 & 57.00 & 10.12 & 69.59 & 52.22 \\
\midrule
DOS (1) & Off & 81.05 & {\bf 41.60} & {\bf 58.54} & {\bf 57.60} & 11.94 & 49.40 & 45.26 \\
DOS (late) & Off & 58.61 & 18.00 & 17.07 & 37.20 & 6.00 & 49.40 & 45.26 \\
Info-Gain & Off & 45.26 & 10.80 & 32.32 & 50.80 & {\bf 17.00} & 70.96 & {\bf 55.24} \\
DAPD & Off & 35.94 & 6.40 & 15.85 & 54.00 & 0.06 & 66.23 & 51.81 \\
\midrule
Left to Right & & 81.05 & {\bf 41.60} & {\bf 58.54} & {\bf 57.60} & 11.94 & 49.40 & 45.26 \\
\bottomrule
\end{tabular}
\caption{Evaluation of decoding methods on Dream-v0-Instruct-7B on GSM8K-CoT, MATH500, HumanEval, MBPP, Trip Planning, 9x9 Sudoku, and 3-number Countdown, temperature 0.1, no
Block Diffusion, and the
few-shot settings, as in the
Table~\ref{tab:benchlrllada} caption.}
\label{tab:benchlrdream}
\end{table}

\subsection{Results}

For LLaDA and Dream, in the chain-of-thought tasks (Tables~\ref{tab:benchlrllada} and~\ref{tab:benchlrdream}, left sides),
left-to-right decoding and DOS have an
advantage, except on Dream GSM8K. However, confidence-based methods  are best for LLaDA on code generation and Sudoku (Table~\ref{tab:benchlrllada}), while future-utility methods are best on Trip Planning (Info-Gain) and Countdown (DAPD).  For Dream, the future utility-maximizing Info-Gain
may beat the confidence methods for flexible generation (Table~\ref{tab:benchlrdream}, right side) and
left-to-right is best even for code (Table~\ref{tab:benchlrdream}, middle).
Apart from a marginal 0.23\% gain for DOS (late) on GSM8K, none of the results show a superiority for the dependency-ready
DOS method on any benchmark except when left-to-right decoding is best.

\subsection{Why DOS Decodes Left-to-Right}

DOS maximizes $Gr(j) = \sum_{i=1}^M Attn(j, i)$.
At least at the first layer, RoPE leads to
``decaying inter-token dependency with increasing relative distances''
in the attention scores (see Section 3.4.3 of \citet{roformer}).
This is a property of an upper bound rather than strict monotonicity,
but if it were strictly monotone, it would imply that $Gr(j) > Gr(k)$
when $M < j < k$.
In this case, $Gr(j)$ would be maximized at $j = M + 1$, and the next token
to the right would be decoded by DOS.
Thus, DOS at the first layer would
never escape autoregressive generation.

\begin{figure}[t]
    \centering
    \includegraphics[width=0.8\linewidth]{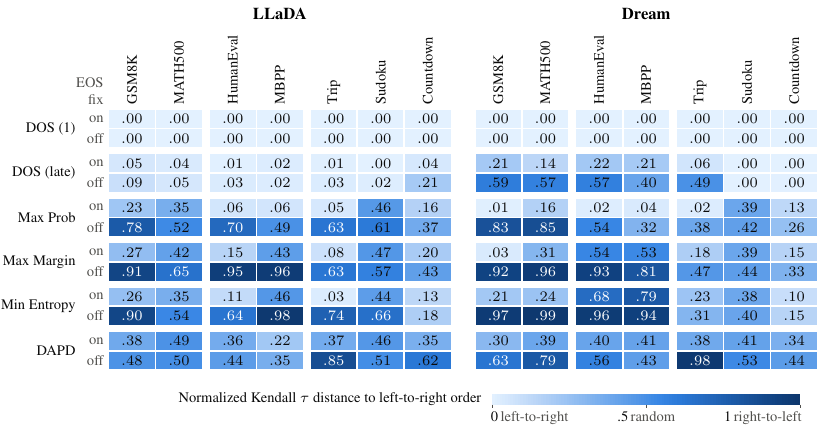}
    \vspace{-2mm}
    \caption{Normalized Kendall $\tau$ distance between each method's decoding order and left-to-right.}
    \label{fig:kendall}
    \vspace{-3mm}
\end{figure}

Figure~\ref{fig:kendall} displays the normalized
Kendall $\tau$ distance \citep{kendalltau} between the
left-to-right decoding
order and the decoding orders chosen by DOS and other
baselines.  This metric counts the fraction of pairs
of position indices $(i, j)$ with $i < j$ such that
position $i$ is decoded after position $j$ ({\em i.e.},
in a different relative order than left-to-right would
decode). At the first layer, DOS stays so close to the left-to-right order that its τ distance rounds to .00 on every task.  Applied to LLaDA, even at a later layer, DOS remains much closer to left-to-right
order than the confidence-based baselines.  
Although DOS escapes left-to-right behavior more on
Dream late layers, it does so without significantly gaining accuracy on
any dataset over using the first layer.
Even with late layers, DOS has nearly zero Kendall $\tau$
on the Sudoku task where dependencies are not expected
to be left-to-right. \rev{In Section~\ref{sec:exp} we further
show that this order makes DOS commit many Sudoku cells before they
are dependency-ready.}

For the other methods, keeping the EOS fix on generally reduces the
$\tau$ distance, consistent with the observation that the EOS is often filled in from the end when these methods are uncontrolled \citep{kim2026rainbow}.  With the EOS fix on, the confidence methods generally have highest $\tau$ for Sudoku, where left-to-right got lower accuracy, and where the notion
of peers suggests a better dependency-ready decoding order.

\subsection{Why Confidence Methods Appeared Not to Work}

\citet{DBLP:conf/acl/ZhouHH26} reported substantial gains
for DOS over maximum probability, maximum margin,
and minimum entropy, but using smaller generation lengths
and not applying an EOS fix.
It is unlikely that EOS is ever the top candidate at the next masked position when decoding from left to right until the response is finished, and we indeed observe that the EOS fix has almost no impact on the DOS results at the first layer.  However, the EOS fix is a very necessary repair for confidence-based strategies, as EOS predictions near the end of the generation limit may be considered confident without any careful analysis.  In our results, we find that simply applying this fix gives some confidence-based methods a double-digit increase in performance, making them outperform DOS with LLaDA on the coding tasks of HumanEval and MBPP where they otherwise would trail it.


\section{Dependency-Ready Decoding Without Attention}
\label{sec:without_attention}

Our motivation (Section~\ref{sec:motivating_dependency})
suggests that dependency-ready decoding should be useful,
but in our experiments, the attention-based DOS fails to
gain a consistent lead
over confidence-based methods or left-to-right decoding,
depending on the task.  In this section, without operationalizing dependency through attention, we establish a theoretical definition based on how much revealing a single
missing token would change the model's prediction.
In this sense, we can examine whether the confidence-based
decoding strategies already consider some form of
dependency readiness. \rev{ Section~\ref{sec:which} then 
measures how often some decoders of Section~\ref{sec:exp} commit dependency-ready tokens.}

\subsection{Defining Unresolved Dependency}
\label{sec:def_dependency}

Let $F_t$ denote the set of positions that remain masked at denoising step $t$,
and let $R_t$ denote the currently revealed context. For a masked target
position $v\in F_t$, the model predicts the categorical distribution $ p_\theta(X_v\mid R_t)$.
Consider another masked position $u\in F_t$, $u\neq v$, and let
$x_u^\star$ denote its ground-truth value. We perform an oracle intervention:
only the missing value at $u$ is revealed, while all other masked positions
remain unchanged. We then measure how much the prediction at $v$ changes using
the Kullback--Leibler divergence
\[
\Delta_t(v\leftarrow u)
=
D_{\mathrm{KL}}
\left(
p_\theta(X_v\mid X_u=x_u^\star,R_t)
\,\middle\|\,
p_\theta(X_v\mid R_t)
\right).
\]

\begin{wrapfigure}[12]{r}{0.35\textwidth}
 \vspace{-15pt}
  \centering
  \includegraphics[width=\linewidth]{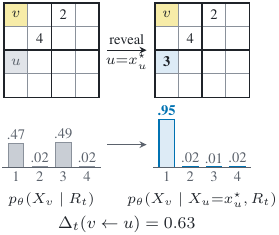}
  \caption{E.g. Revealing $u$ at $x^\star_u = 3$ rules out 3 for $v$, so the top-1 prediction at $v$ flips from 3 to 1. }
  \label{fig:delta}
\end{wrapfigure}

A large value of $\Delta_t(v\leftarrow u)$ indicates that the prediction at
$v$ is sensitive to the information at $u$: revealing $u$ substantially changes the model's belief about $v$, possibly
enough to introduce a \emph{top-1 prediction flip}, as in Figure~\ref{fig:delta}.
Conversely, a small value indicates that revealing $u$ has little effect on the current prediction.

A target prediction may depend on multiple unresolved positions. We therefore
summarize its strongest unresolved dependency as
\[
\mu_t(v)
=
\max_{\substack{u\in F_t\\u\neq v}}
\Delta_t(v\leftarrow u).
\]

We refer to $\mu_t(v)$ as the \emph{Pre-Commit score}. A small
$\mu_t(v)$ means that the prediction at $v$ is relatively stable to future
revelations, whereas a large value indicates that the prediction remains exposed
to information that has not yet been resolved.

The ground-truth intervention is deliberate. Revealing the model's own current
prediction at $u$ would measure the effect of confirming its existing belief,
rather than the effect of receiving the information that is actually missing.
In particular, if the model were confidently wrong about $u$, self-conditioning
could reinforce the incorrect hypothesis and conceal a genuine dependency.
Ground-truth conditioning instead measures the change in the
prediction at $v$ once the correct missing information becomes available.

\subsection{\rev{Large margins compared to the Pre-commit score certify dependency readiness}}\label{sec:which0}

Let $q_{t,v} (x)$ denote the probability with which token $x$ is predicted
to fill position $v$ at time slot $t$, and $p_{t,v}^{(u)} (x)$ denote the
same probability if the ground truth token at $u$ is also revealed, \rev{$i.e., q_{t,v}(x) = p_\theta(X_v = x \mid \mathcal{R}_t)
\quad \text{and} \quad
p_{t,v}^{(u)}(x) = p_\theta(X_v = x \mid X_u = x_u^*, \mathcal{R}_t)$}.
In this notation, $
\mu_t(v) =\max_{\substack{u\in F_t\\u\neq v}} D_{KL} (p_{t,v}^{(u)} \vert\vert q_{t,v})
\ldotp $

Let $w_1$ and $w_2$ be the top two tokens predicted to fill position $v$
by $q_{t,v}$, and $m_t (v) = q_{t,v} (w_1) - q_{t,v} (w_2)$ be the margin.
We can show that if $m_t (v)$ is sufficiently large \rev{relative to $\mu_t(v)$}, then the decoding of
$w_1$ by Maximum-Margin is not premature, in the sense that if another
token were revealed, $w_1$ would remain the top-1
prediction at v.

{\bf Proposition.} If $m_t(v) > \sqrt{2\mu_t(v)}$, then $p_{t,v}^{(u)}(w_1) > 
p_{t,v}^{(u)}(x)$ for any masked $u$ and any $x \neq w_1$.

{\em Proof.} Let $\norm{p_{t,v}^{(u)} - q_{t,v}}_\infty = \sup_{A} \abs{p_{t,v}^{(u)}(A) - q_{t,v}(A)}$, where $A$ denotes a measurable
event. Then
\begin{align}
p_{t,v}^{(u)}(w_1) - p_{t,v}^{(u)}(x)
&= q_{t,v}(w_1) - q_{t,v}(x) + \big(p_{t,v}^{(u)}(w_1) - q_{t,v}(w_1)\big) - \big(p_{t,v}^{(u)}(x) - q_{t,v}(x)\big) \nonumber \\
&\geq m_t(v) - 2\norm{p_{t,v}^{(u)} - q_{t,v}}_\infty \;\geq\; m_t(v) - \sqrt{2\mu_t(v)} \;
>\; 0,
\end{align}

where the second inequality is Pinsker's inequality \citep{pinsker}, $\norm{p_{t,v}^{(u)} - q_{t,v}}_\infty \leq \sqrt{\mu_t(v)/2}$, and the last follows from our hypothesis. \qed

In this way,  a margin that is large relative to the Pre-Commit score implies that the revelation of another
masked token will not affect the token selection.

\begin{figure}[htbp]
\centering
\begin{tikzpicture}
\begin{axis}[
  width=0.9\linewidth, height=4.2cm,
  scale only axis=false,
  xmin=0.5, xmax=14.5,
  ymin=-2, ymax=52,
  ytick={0,20,40},
  xtick={1,...,14},
  xticklabels={$-2$, $-1.5$, $-1.25$, $-1$, $-0.75$, $-0.5$, $-0.25$,
               $-0.1$, $-0.05$, $0$, $0.25$, $0.5$, $0.75$, $1$},
  xticklabel style={font=\tiny},
  yticklabel style={font=\tiny},
  ylabel={Flip rate (\%)},
  ylabel style={font=\scriptsize, yshift=-4pt},
  xlabel={Upper edge of the block of $d(v)$},
  xlabel style={font=\scriptsize, yshift=3pt},
  legend style={at={(0.99,0.55)}, anchor=east, font=\tiny,
                draw=none, fill=none, row sep=-2pt},
  legend cell align=left,
  unbounded coords=jump,
  axis on top,
]

\fill[red!12]    (axis cs:0.5,-2)  rectangle (axis cs:7.5,52);
\fill[orange!15] (axis cs:7.5,-2)  rectangle (axis cs:10.5,52);
\fill[green!12]  (axis cs:10.5,-2) rectangle (axis cs:14.5,52);
\draw[dotted, thick] (axis cs:10.5,-2) -- (axis cs:10.5,52);

\node[font=\tiny\bfseries] at (axis cs:4,47)    {No guarantee};
\node[font=\tiny\bfseries] at (axis cs:9,47)    {Near boundary};
\node[font=\tiny\bfseries] at (axis cs:12.5,47) {Stability region};

\addplot[semithick, black, mark=*, mark size=1.4pt, xshift=-2pt] coordinates {
  (1,40.0) (2,39.3) (3,41.0) (4,36.7) (5,34.0) (6,19.0) (7,6.4)
  (8,3.4) (9,1.2) (10,0.21) (11,0) (12,0) (13,0) (14,0)
};
\addlegendentry{$4\times4$ Sudoku}

\addplot[semithick, dashed, blue!70!black, mark=square*, mark size=1.7pt,
         mark options={solid, fill=white}, xshift=2pt] coordinates {
  (1,5.6) (2,14.3) (3,22.3) (4,40.9) (5,46.5) (6,23.3) (7,16.9)
  (8,0.82) (9,nan) (10,nan) (11,0) (12,0) (13,nan) (14,nan)
};
\addlegendentry{$9\times9$ Sudoku}

\end{axis}
\end{tikzpicture}
\vspace{-7pt}
\caption{
Top-1 flip rate per block of $d(v) = m(v) - \sqrt{2\mu(v)}$ (\texttt{Max-Margin}, exact sweeps, LLaDA-8B-Instruct). On $4\times4$, the rate falls from about $40\%$ to $0.21\%$
just below the boundary. On $9\times9$, it peaks at $46.5\%$. Neither board has a flip above
the boundary ($0$ in $3{,}777$ and $467$ reveals). 
}
\label{fig:flip_rate_boundary}
\end{figure}
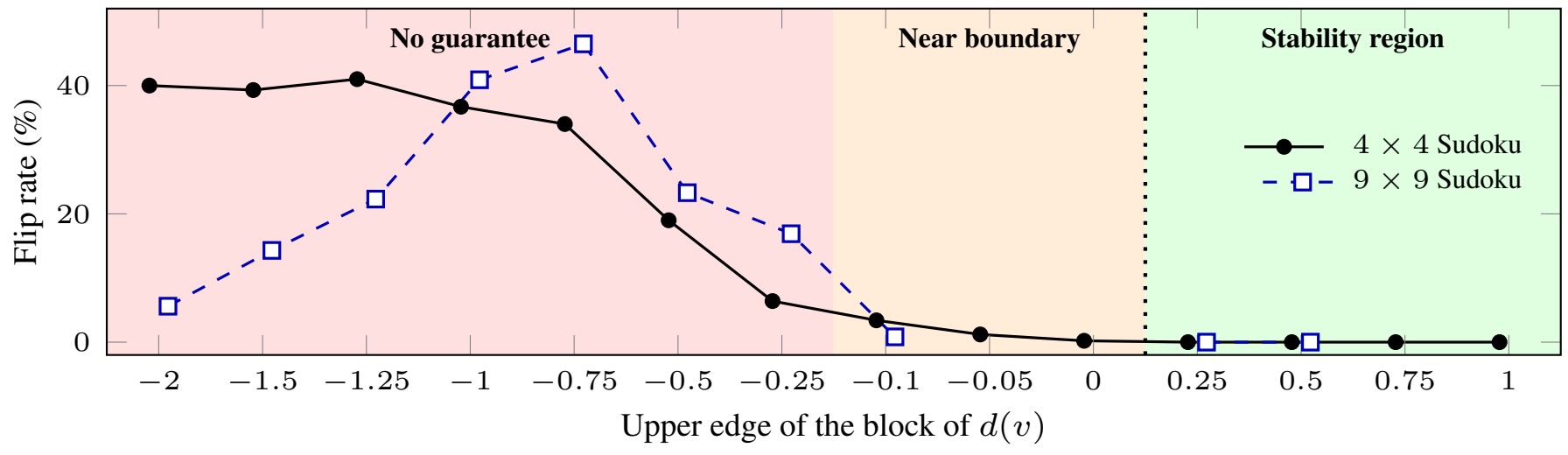



\rev{The proposition extends to revealing any set $U\subseteq F_t\setminus\{v\}$
of $k$ tokens. Define
\[
\Delta_t(v\leftarrow U):=D_{\mathrm{KL}}\!\left(p_\theta(X_v\mid X_U=x_U^\star,R_t)\,\middle\|\,p_\theta(X_v\mid R_t)\right),
\;\;\;
\mu_t^k(v):=\max_{\substack{U\subseteq F_t\setminus\{v\}\\|U|=k}}\Delta_t(v\leftarrow U).
\]
By the same argument, $m_t(v)>\sqrt{2\mu_t^k(v)}$ guarantees that
revealing any $k$ masked tokens does not change the top-1 prediction.
Larger subsets yield a stricter margin criterion.
}
\paragraph{Empirical validation.}
We test this stability condition empirically on 4$\times$4 and 9$\times$9 Sudoku with the  LLaDA Instruct model. At every denoising step $t$ and for every masked position $v$, we compute the margin $m_t(v)$ and $\mu_t(v)$. To compute $\mu_t(v)$, we reveal each other masked position $u$ at its ground-truth token, one at a time, and check whether the top-1 prediction at $v$ changes. As guaranteed by the proposition, the top-1 prediction never changes when the condition holds (0 of 3{,}777 revelations on 4$\times$4 and 0 of 467 on 9$\times$9).
Far below the stability boundary, by contrast, top-1 changes are frequent (Figure~\ref{fig:flip_rate_boundary}): for all bins with $m_t(v)-\sqrt{2\mu_t(v)}\leq -0.75$, 34--41\% of revelations change the top-1 prediction on 4$\times$4, while on 9$\times$9 the rate reaches 46.5\%, nearly half.  We note that these low-margin token positions may be unlikely to be chosen by Maximum-Margin decoding.  However, the results suggest that the proposition leaves room for future work on a non-oracle dependency-ready decoding method, likely one that does not rely on attention, to identify dependency-ready positions that maximum margin may fail to select when the margin is small relative to $\sqrt{2\mu_t(v)}$.


\definecolor{cMM}{HTML}{0000FF}\definecolor{cMP}{HTML}{9C5A1A}
\definecolor{cME}{HTML}{000000}\definecolor{cDOS}{HTML}{FF0000}
\pgfplotsset{marginbins/.style={
  width=\linewidth, height=3.0cm, scale only axis=false,
  symbolic x coords={b1,b2,b3,b4,b5,b6}, xtick=data,
  xticklabels={0--.1,.1--.3,.3--.5,.5--.7,.7--.9,.9--1},
  enlarge x limits=0.06,
  tick label style={font=\tiny}, label style={font=\scriptsize},
  xlabel={Margin $m_t(v)$}, xlabel style={yshift=4pt}, ylabel style={yshift=-6pt},
  ymajorgrids, grid style={black!12}, axis lines*=left,
  every axis plot/.append style={line width=0.7pt, mark size=1.3pt},
}}

\subsection{\rev{Which decoders commit dependency-ready tokens?}}\label{sec:which}

\begin{figure}[t]
\centering
\begin{minipage}[b]{0.33\linewidth}
\centering\scriptsize\setlength{\tabcolsep}{1.3pt}
\begin{tabular}{lrrrrrr}
\toprule
& \multicolumn{2}{c}{Guar.} & \multicolumn{2}{c}{Stable} & \multicolumn{2}{c}{Unstable}\\
\cmidrule(lr){2-3}\cmidrule(lr){4-5}\cmidrule(lr){6-7}
 & \% & Acc. & \% & Acc. & \% & Acc.\\
\midrule
Max Prob.   & 61.8 & 77.3 & 12.9 & 63.2 & 25.3 & 8.9 \\
Max Marg. & 62.9 & 76.8 & 13.3 & 56.9 & 23.9 & 7.5 \\
Min Ent.   & 59.7 & 77.3 & 11.6 & 61.2 & 28.7 & 12.2\\
L2R / DOS  & 47.4 & 73.1 & 12.1 & 70.4 & 40.5 & 20.7\\
\bottomrule
\end{tabular}
\vspace{1pt}
\captionof{table}{\rev{Stability tiers. \%: share of commits; Acc.: \% correct in the tier.}}
\label{tab:tiers}
\end{minipage}\hfill
\begin{minipage}[b]{0.63\linewidth}
\captionsetup{type=figure}\centering
\begin{tikzpicture}
\begin{axis}[hide axis, scale only axis, xmin=0, xmax=1, ymin=0, ymax=1, height=1pt, width=\linewidth,
  legend columns=4, legend style={font=\tiny, draw=none, column sep=4pt, at={(0.5,0.5)}, anchor=center},
  every axis plot/.append style={line width=0.7pt, mark size=1.3pt}]
\addlegendimage{cMM, solid, mark=*}\addlegendentry{Max Margin}
\addlegendimage{cMP, densely dashed, mark=triangle*}\addlegendentry{Max Prob}
\addlegendimage{cME, densely dashed, mark=diamond*}\addlegendentry{Min Entropy}
\addlegendimage{cDOS, solid, mark=square*}\addlegendentry{L2R / DOS}
\end{axis}
\end{tikzpicture}\\[-4pt]
\begin{subfigure}[b]{0.5\linewidth}
\centering
\begin{tikzpicture}
\begin{axis}[marginbins, ymin=-2, ymax=50, ytick={0,20,40}, ylabel={Commits (\%)}]
\addplot[cMM, solid, mark=*] coordinates {(b1,0.1) (b2,0.3) (b3,1.8) (b4,15.0) (b5,38.4) (b6,44.3)};
\addplot[cMP, densely dashed, mark=triangle*] coordinates {(b1,0.2) (b2,0.4) (b3,3.0) (b4,17.2) (b5,36.2) (b6,43.0)};
\addplot[cME, densely dashed, mark=diamond*, mark size=1.7pt] coordinates {(b1,1.1) (b2,2.7) (b3,5.2) (b4,13.8) (b5,35.0) (b6,42.3)};
\addplot[cDOS, solid, mark=square*] coordinates {(b1,9.3) (b2,16.4) (b3,12.6) (b4,11.9) (b5,14.4) (b6,35.4)};
\end{axis}
\end{tikzpicture}
\vspace{-12pt}
\caption{\scriptsize Where decoders commit.}
\label{fig:share}
\end{subfigure}\hfill
\begin{subfigure}[b]{0.5\linewidth}
\centering
\begin{tikzpicture}
\begin{axis}[marginbins, ymin=-4, ymax=104, ytick={0,50,100}, ylabel={Flip rate (\%)}]
\addplot[cMM, solid, mark=*] coordinates {(b1,0.0) (b2,28.6) (b3,45.6) (b4,36.7) (b5,25.9) (b6,16.8)};
\addplot[cMP, densely dashed, mark=triangle*] coordinates {(b1,0.0) (b2,32.3) (b3,50.2) (b4,39.8) (b5,26.5) (b6,16.8)};
\addplot[cME, densely dashed, mark=diamond*, mark size=1.7pt] coordinates {(b1,88.5) (b2,72.3) (b3,58.0) (b4,43.5) (b5,27.6) (b6,16.8)};
\addplot[cDOS, solid, mark=square*] coordinates {(b1,98.4) (b2,84.1) (b3,59.7) (b4,41.4) (b5,26.3) (b6,3.8)};
\addplot[only marks, mark=*, mark size=1.3pt, mark options={draw=cMM, fill=white, solid}] coordinates {(b1,0.0) (b2,28.6)};
\addplot[only marks, mark=triangle*, mark size=1.3pt, mark options={draw=cMP, fill=white, solid}] coordinates {(b1,0.0)};
\end{axis}
\end{tikzpicture}
\vspace{-12pt}
\caption{\scriptsize How often they flip.}
\label{fig:flipmargin}
\end{subfigure}
\vspace{-8pt}
\caption{\rev{Commits by margin: (a) share of each decoder's commits; (b) share of those flipped by a single ground-truth revelation. Hollow: bins with $<$30 commits.}}
\label{fig:margin}
\end{minipage}
\vspace{-3mm}
\end{figure}

\rev{
We decode 100 $9\times9$ Sudoku puzzles with Dream, one token per step
with the EOS fix off (Table~\ref{tab:tiers}, Figure~\ref{fig:margin}). For each
commit at $v$, we reveal every other masked position, one at a time, at its
ground-truth value. A commit is \emph{guaranteed} if $m_t(v) > \sqrt{2\mu_t(v)}$,
\emph{stable} if the criterion fails but no revelation flips the top-1 prediction,
and \emph{unstable} otherwise. On Dream, DOS~(1) decodes exactly left-to-right
(Figure~\ref{fig:kendall}), so the two share a row.
}

\rev{
The confidence-based methods decode tokens that are overturned less often than DOS:
63\% of Maximum-Margin commits meet the criterion and only 24\% are overturned by a single
revelation (Table~\ref{tab:tiers}), with similar proportions for Maximum-Probability and Minimum-Entropy.
In contrast, more than 40\% of DOS commits are
unstable.
}

\rev{We note that
stable or guaranteed commits are far more often
 correct: 56--77\%, against 8--21\% of unstable ones
 (Table~\ref{tab:tiers}), and unstable commits account for more than half of each
 decoder's errors. A prediction satisfying the criterion can still be wrong, with probability 22.7--26.9\%, giving an empirical upper bound on the frequency with which more than one token revelation might be necessary in order to predict the correct token.}

\rev{
Figure~\ref{fig:margin} distinguishes high margin from criterion satisfaction.  Generally, high margin predictions are less likely to be overturned by a revelation.  Maximum-Margin and Maximum-Probability make occasional low-margin predictions without a flip, but DOS makes such predictions more often and they are more likely to be overturned.}



\section{Conclusion}

\rev{Our results separate the principle of dependency-ready decoding from its attention-based implementation.}
Empirically, we generally did not observe that an attention implementation
of dependency-ready decoding (DOS) achieved gains over
confidence-based
or future utility strategies, except on benchmarks where
left-to-right decoding was beneficial.  We were further
discouraged by the left-to-right
behavior which resulted from the use of attention as a proxy for
dependency at the first layer, even on benchmarks like Sudoku
where other kinds of dependencies were expected.  

\rev{Without resorting to attention, we offered a theoretical definition of what dependency-readiness could mean, in terms of making predictions not overturned by  revealing one or a set of
tokens, and defined the Pre-Commit score as a measure of a token position's prematurity in decoding.
We showed theoretically
that Maximum-Margin decoding already follows this principle
whenever the margin is large enough compared to the Pre-Commit score.
Empirically, we also showed that on Sudoku, Maximum-Margin predictions could be overturned by token revelations less often than DOS predictions, and that decoding in positions not satisfying the dependency-ready criterion more often led to errors.}

\rev{To offer an improvement over existing techniques, we suggest that future dependency-ready decoders should address the situation where the margin-based
 guarantee does not apply, and should prioritize decoding positions without relying on attention.}

\newpage

\section{Reproducibility Statement}
We made several efforts to ensure our results can be reproduced. Our experimental setup describes the models, benchmarks, evaluation harness, EOS Fix, and hardware, as well as the configuration of each decoding strategy, including the attention layers used for DOS (1) and DOS (late). The prompts follow those provided in the public 
Dream repository\footnote{\url{https://github.com/DreamLM/Dream}}, 
and the few-shot settings and generation lengths are reported in the setup and with the main results. We ran DAPD and Info-Gain Sampler with the official code released by their authors, and since no official code was available for DOS, we reimplemented it following 
\citet{DBLP:conf/acl/ZhouHH26}, 
using the groundedness score defined earlier. All methods share the same prompts, model weights, and decoding budget and reveal one token per step without blocks, so they differ only in how they choose which positions to reveal. LLaDA is decoded with temperature 0, and Dream with temperature 0.1, following the LLaDA and Dream papers. We also give the full setup of our motivating experiment and the definition of the Kendall $\tau$ distance we use to compare decoding orders. Finally, we define the Pre-Commit score and the oracle intervention, state the assumptions of the Proposition with its complete proof, and describe the stability check we ran.

\section{Ethics Statement}
All authors adhere to the ICLR Code of Ethics. We study decoding strategies for public MDMs on public benchmarks, involving no human subjects or personal data. All models and datasets follow their original licenses. 
No new capabilities or risks beyond those of the underlying models and pre-existing decoding methods are introduced.

\section{AI Usage Statement}
In this work, we used generative AI tools to help refine our hypotheses and experimental design and to help implement our methods, including the decoding strategies and the Pre-Commit score computations. We have not used generative AI tools to  interpret results or clean and reformat datasets.  Generating synthetic data sets, translation, and qualitative or thematic data analysis are not applicable to this work. Additionally, we used generative AI tools to help write and edit other code, such as evaluation and plotting scripts, to help create figures, to draft and edit parts of the paper, and to search for relevant literature. We mainly used Claude Code to refine code, OpenAI's GPT models to rephrase and edit text and to search papers. All AI-assisted code was checked and verified by the authors.  The definitions of Pre-Commit score and the oracle intervention were developed without AI. The Proposition was suggested by GPT-6 Astra but its proof was written by the authors. All citations were written and checked manually by the authors. 

\newpage

\bibliography{iclr2027_conference}
\bibliographystyle{iclr2027_conference}

\end{document}